\documentclass{article}

\PassOptionsToPackage{numbers, compress}{natbib}
\usepackage[preprint]{neurips_2021}

\usepackage[accsupp]{axessibility}  
\usepackage[utf8]{inputenc} 
\usepackage[T1]{fontenc}    
\usepackage{hyperref}       
\usepackage{url}            
\usepackage{booktabs}       
\usepackage{amsfonts}       
\usepackage{nicefrac}       
\usepackage{microtype}      
\usepackage{xcolor}         
\usepackage{xfrac}
\usepackage{mathtools}
\usepackage{csquotes}
\usepackage{multicol}
\usepackage{multirow}
\usepackage{tikz}
\usetikzlibrary{bayesnet}

\author{
Nathan Drenkow$^{1,2}$,
Numair Sani$^1$,
Ilya Shpitser$^1$,
Mathias Unberath$^1$ \\
\\ [0.25cm]
The Johns Hopkins University \\
$^1$Department of Computer Science~~$^2$Applied Physics Laboratory
}

\begin{document}

\title{A Systematic Review of Robustness in Deep Learning for Computer Vision: Mind the gap?}

\maketitle

\begin{abstract}
Deep neural networks for computer vision are deployed in increasingly safety-critical and socially-impactful applications, motivating the need to close the gap in model performance under varied, naturally occurring imaging conditions. Robustness, ambiguously used in multiple contexts including adversarial machine learning, refers here to preserving model performance under naturally-induced image corruptions or alterations.
  
We perform a systematic review to identify, analyze, and summarize current definitions and progress towards non-adversarial robustness in deep learning for computer vision. We find this area of research has received disproportionately less attention relative to adversarial machine learning, yet a significant robustness gap exists that manifests in performance degradation similar in magnitude to adversarial conditions. 
  
Toward developing a more transparent definition of robustness, we provide a conceptual framework based on a structural causal model of the data generating process and interpret non-adversarial robustness as pertaining to a model's behavior on corrupted images corresponding to low-probability samples from the unaltered data distribution. We identify key architecture-, data augmentation-, and optimization tactics for improving neural network robustness. This robustness perspective reveals that common practices in the literature correspond to causal concepts. We offer perspectives on how future research may mind this evident and significant non-adversarial robustness gap. 
  
\end{abstract}

\section{Introduction}
\label{sec:introduction}

The considerable progress made by deep neural networks (DNNs) for vision tasks has been closely paralleled by discoveries of susceptibility to image-level alterations including both adversarial and natural manipulations.  Initial findings~\cite{szegedy2013intriguing} demonstrated that imperceptible perturbations to images led to a change in prediction by the model from confidently correct to confidently incorrect.  Follow-up works have demonstrated DNNs' susceptibility to natural, non-adversarial corruptions~\cite{hendrycks2019_benchmarking, laugros2019_are-adversarial} which represent conditions more likely to occur in the real world compared to adversarial perturbations and pose equally serious threats to DNN performance. Significant gaps in DNN performance are consistently observed between evaluations on clean and naturally degraded imaging conditions (often as much as 30-40\% accuracy decreases), and these findings raise concerns about the reliability of DNNs as they are integrated into systems with increasingly high safety and social stakes.  The significant vulnerability of models to infrequent yet natural image corruptions suggests a need to re-prioritize our understanding of model performance on natural data before focusing on resilience to adversarial attack scenarios.

The research community has responded to these findings by developing an array of new approaches to characterize and address model ``robustness''. Several years into this pursuit, it's time to review proposed solutions to these problems and assess what measurable progress has been made, what problems remain, and potential directions for future research. This systematic review article aims to examine how non-adversarial robustness itself is defined and then to identify, organize, and analyze key datasets, algorithms, and evaluation protocols for characterizing and improving DNN robustness in that context. In short, beyond adversarial machine learning, what is non-adversarial robustness, is there a robustness gap in deep learning, and to what extent should we mind it?

\section{Background and Definitions}
\subsection{Deep Neural Networks}
We focus here on robustness in deep learning and the use of deep neural networks for computer vision tasks.  In the most general form, we consider a deep neural network as a composition of functions or computational layers which map from input space in the image domain to a prediction.  These networks are highly parameterized, requiring large datasets and/or sophisticated data augmentation/generation schemes in combination with various forms of parameter optimization (often stochastic gradient descent). While the origins of their susceptibility are not fully understood (a key motivation for this review), over-parameterization, compact training domains, and under-regularized optimization are contributing factors towards DNNs' inability maintain their performance in the face of small, natural perturbations of the inputs.

\subsection{Defining Robustness}
We aim to study various definitions and forms of robustness found in the literature and to build a more holistic view of the existing deep learning research challenges beyond achieving state-of-the-art performance on common vision benchmark tasks.  

\subsubsection{Common Interpretations}
In general, \emph{robustness} is an overloaded term and has taken on a range of interpretations in the computer vision community including, but not limited to, raw task performance on held-out test sets, maintaining task performance on manipulated/modified inputs, generalization within/across domains, and resistance to adversarial attacks. While these are all desirable robustness objectives for deep learning models to achieve, clear and formal definitions are lacking outside of the adversarial domain which relies on well-defined constraints such as $L_p$-norm bounds on adversarial perturbations. While one core aim of this systematic review is to identify specific ways in which non-adversarial ``robustness'' is defined in the literature, we establish a reference point for reviewing any such definitions using the concepts underlying causal inference.

\subsubsection{A Causal Perspective on Defining Robustness}
\label{subsub:robust_defn}

We present here a causal view of robustness to provide a common \textit{conceptual framework} for understanding various definitions/perspectives of robustness in the literature given that our review found no suitable alternatives. 

Fundamentally, robustness is a relative, rather than absolute, measure of model performance. We seek to consider three key elements of robustness: (1) the image alteration or corruption itself, (2) the design/optimization of the model to mitigate against this corruption, and (3) the nature of the evaluation and measures of performance. In particular, the form and properties of (1) often drive the design choices of (2) and (3), and as such, must be used to limit the scope of our review.

\begin{figure*}[h!]
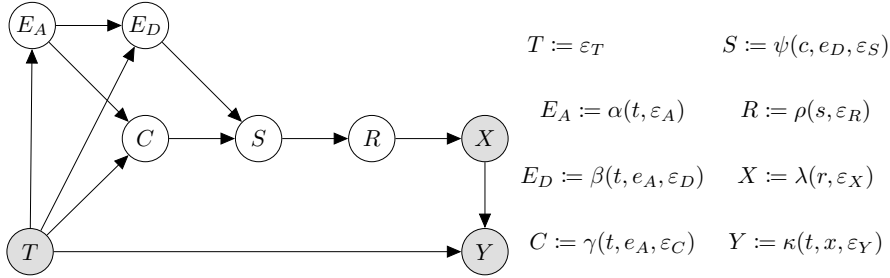

  \centering
  \resizebox{12cm}{!}{
  \tikz{ %
    \node[latent] (env_concept) {$E_D$} ; %
    \node[latent, left=of env_concept] (env_appearance) {$E_A$} ; %
    \node[latent, below=of env_concept] (concept) {$C$} ; %
    \node[latent, right=of concept] (sensor) {$S$} ; %
    \node[obs, below=of concept, xshift=-1.75cm] (task) {$T$} ; %
    \node[latent, right=of sensor] (render) {$R$} ; %
    \node[obs, right=of render] (x) {$X$} ; %
    \node[obs, below=of x] (y) {$Y$} ; %
    
    \node[above right of= x, xshift = 15pt, yshift = 20pt](eq_T){$T \coloneqq \varepsilon_T$};
    \node[below of= eq_T, xshift = 20pt](eq_ea){$E_A \coloneqq \alpha(t, \varepsilon_A)$};
    \node[below of= eq_ea](eq_ec){$E_D \coloneqq \beta(t, e_A, \varepsilon_D)$};
    \node[below of= eq_ec](eq_C){$C \coloneqq \gamma(t, e_A, \varepsilon_C)$};
    \node[right of= eq_T, xshift = 75pt](eq_S){$S \coloneqq \psi(c, e_D, \varepsilon_S)$};
    \node[below of= eq_S](eq_R){$R \coloneqq \rho(s, \varepsilon_R)$};
    \node[below of= eq_R](eq_X){$X \coloneqq \lambda(r, \varepsilon_X)$};
    \node[below of= eq_X](eq_Y){$Y \coloneqq \kappa(t, x, \varepsilon_Y)$};
    
    \edge {task} {env_concept} ; %
    \edge {task} {env_appearance} ; %
    \edge {task} {concept} ; %
    \edge {env_appearance} {concept} ; %
    \edge {env_appearance} {env_concept} ; %
    \edge {env_concept} {sensor} ; %
    \edge {concept} {sensor} ; %
    \edge {sensor} {render} ; %
    \edge {render} {x} ; %
    \edge {x} {y} ; %
    \edge {task} {y} ; %
  }
  }
  \caption{The data generating process is described by an SCM with its corresponding DAG above. The task ($T$) separates concepts of interest ($C$) from environment distractor concepts ($E_D$) and appearance nuisance factors ($E_A$). Scenes composed by sampling from these variables are captured by a sensor a sensor ($S$) while a renderer ($R$) may make additional alterations before yielding the final image ($X$). Labels ($Y$) are generated given both the task and the image. Non-adversarial robustness is viewed as dealing with images appearing as corrupted, altered, or otherwise transformed and which correspond to samples from the unaltered distribution where values of $E, S, R$ are low-probability.
  }
  \label{fig:scm}
\end{figure*}

We consider robustness in the context of deep learning for computer vision through a causal lens. We assume the Data Generating Process (DGP) can be represented by a Structural Causal Model (SCM)~\cite{pearl2009causality,bollen1989structural, peters2017elements} with its corresponding Directed Acyclic Graph (DAG) and its SCM. Knowledge of scene construction, physics, and other aspects of the image generation process naturally allow the specification of such a model.  As different imaging domains implement different generation processes, the causal approach allows researchers to naturally state assumptions and define appropriate priors through causal models tied to their application.  For the purposes of this review, we present a general DGP in Figure~\ref{fig:scm} which is sufficient to cover a wide range of imaging scenarios.

\paragraph*{\textbf{Common DGP}}
In this model, $T$ is defined as the task (i.\,e, the set of all image-related tasks/questions) which separates a set of concepts $C$ (e.\,g., classes of interest) from their environment $E$ (e.\,g., appearance, background, and nuisance factors $E_A$, distractor concepts, $E_D$). Examples of $E_A$ might include weather, lighting, or other atmospheric conditions while $E_D$ might include other objects/concepts in the scene which occlude or distract from classes of interest.  Scenes composed by sampling from these variables are captured by a sensor $S$ which maps from physical reality to the image domain.  Sensor properties may include noise characteristics, pose, and/or lens properties/distortion.  An additional rendering step $R$ may modify appearances or image statistics, such as compressing a raw image (e.\,g., JPEG), performing quantization, converting it to a stylized counterpart, or leaving the raw image unaltered. Image $X \in R^d$ is the output of the renderer and is used, in combination with $T$, to determine the label $Y$. For additional examples of corruptions and their relation to this SCM see Table~\ref{tab:corruptions}.

The SCM represents causal relationships among these variables using \emph{structural equations} or causal mechanisms, which determine the value of each variable in terms of its parents in the graph, as well as an exogenous noise term (denoted by $\varepsilon$).
Figure~\ref{fig:scm} induces the following structural equations:
$\alpha: T \times \mathcal{E}_A \rightarrow \mathbb{R}^{E_A}$, $\beta: T \times E_A \times \mathcal{E}_D \rightarrow \mathbb{R}^{E_D}$, $\gamma: T \times E_A \times \mathcal{E}_c \rightarrow \mathbb{R}^C$, $\psi: C \times E_D \times \mathcal{E}_S \rightarrow \mathbb{R}^S, \rho: S \times \mathcal{E_R} \rightarrow \mathbb{R}^R$, $\lambda: R \times \mathcal{E}_X \rightarrow \mathbb{R}^X$ and $\kappa: T \times X \times \mathcal{E}_Y \rightarrow \mathcal{Y}$.

Unlike statistical graphical models, causal models, such as the SCM, permit representing and reasoning about counterfactual situations where \emph{intervention operations} replace structural equations associated with specific variables with other structural equations.  For example, the structural equation $\gamma: \mathcal{T} \times E_A \times \mathcal{E_D} \rightarrow \mathbb{R}^C$ may be replaced by $\gamma^*$ that always outputs a constant $c$.  Such an intervention, written as
$\text{do}(c)$ in \cite{pearl2009causality}, represents a modified data generating process where only a specific concept $c$ is used for all causally-subsequent variables, such as the sensor or the renderer in this case.  Importantly, variables causally prior to $C$, such as the task or the environment, remain unchanged by the intervention.  Thus, interventions differ in this crucial respect from the more familiar conditioning operation.  While interventions that set variables to constants are most common in causal inference, \emph{soft interventions} \cite{eberhardt14soft} have also been considered.  One example of such an intervention sets a given variable $E_A$ to respond according to a user specified distribution $p^*(E_A)$.

Underlying the SCM is a joint distribution over all causal variables and whose conditional probabilities are faithful to the structure defined by the DAG. Furthermore, such a model allows us to explicitly state our assumptions about the image generation process and the likelihood of various imaging conditions and scenarios for the underlying distribution. 
While the true marginal/joint distributions cannot be known in most situations, initial distributional assumptions can be validated or modified with the careful collection and analysis of additional metadata. This could include the time of day, location, sensor calibration parameters, or other scene-related information, all of which can be related to specific nodes in the causal graph and might be used to infer/quantify causal relationships. 

Finally, once the SCM of the DGP is defined, we can consider the distribution as stationary under those assumptions, thus allowing for downstream analysis such as estimating sample likelihood or alignment of train/validation/test \textit{sample} distributions with the assumed distribution underlying the DGP. Practically speaking, if the assumed DGP allows for naturally corrupted images with non-zero probability, then even if the train/validation/test sets do not contain naturally corrupted images, it does not change the original DGP's distributional assumptions.  However, if a separate scenario, the DGP is assumed to assign zero probability to naturally corrupted images, then constructing a test set with altered images would constitute a distribution shift. The key point here is that the causal perspective provides a means for stating distributional assumptions such that we can study robustness in specific and targeted cases.

\paragraph*{\bf Causal Framework for Robustness}

We consider non-adversarial robustness as dealing with a model's behavior on altered/transformed images corresponding to samples from the underlying distribution of the DGP where values for environmental, sensor, or rendering conditions (i.e., $E, S, R$) are low-probability. For instance, values $e_{corrupt}$ of the environmental nuisance factors $E_A$ such that $p(E_A = e_{corrupt}) < \delta$ for a small $\delta$ representing the ``tail'' of the marginal $p(E_A)$ where an example of such a value of $e_{corrupt}$ might correspond to a rare weather event or lighting condition. Since the variables $\{E_A,E_D,C,S\}$ capture high-level properties/measurements/configurations of the real world, this long-tail interpretation is consistent with the observation that both frequent and rare phenomena exist naturally in the physical world. 

Predictive model performance is commonly measured in terms of expected loss (with respect to the observed data distribution).  Thus, a model can often achieve seemingly high performance without properly taking into account imaging conditions that occur rarely, but are nevertheless important.  By contrast, a model is robust if it is able to properly deal with many kinds of images, corresponding to both common and rare phenomena whose likelihoods are well-defined in the context of the SCM. By considering robustness as dealing with samples from the tail of the true distribution (i.e., where the likelihood of the sample is smaller than some threshold), the emphasis is placed on studying imaging conditions which are sufficiently rare that they are not likely to be found in the training sample distribution where a DNN could be more directly optimized against these conditions.

Thus, a natural approach for evaluating model robustness is to consider how the model performs on data obtained from counterfactually-altered versions of Fig. \ref{fig:scm}, where rare phenomena are artificially emphasized via soft interventions.  For example, performing an intervention on an environmental nuisance variable in $E_A$ and setting its value corresponding to these rare phenomena (e.\,g., extreme weather), will result in an image $X$ which appears measurably different from the case where $E_A$ is left unaltered, and most of its values correspond to high probability events (e.\,g. normal weather).

Under this causal perspective, distribution shifts are identified as a new realization of the underlying distribution given the SCM whereas non-adversarial image corruptions/alterations are distinguished as low probability samples from the \emph{unaltered} original distribution. While robustness evaluations employ interventions on specific variables in Fig.~\ref{fig:scm} to target specific low-probability events, distribution shifts are viewed as \emph{true} changes to distributions of one or more variables in the SCM. 

For instance, in a street sign classification task for an autonomous vehicle, consider the case where the initial DGP assumes street sign classes ($C$), weather ($E_A$), other occluder/distractor concepts ($E_D$) are drawn from a single geographic location. Then, we can see that changing the geographical location/country where images are collected defines a clear distribution shift since the joint density $p(E_A, E_D, C, X, Y)$ changes as a result of, e.\,g., new geographical features, sign concepts and appearances, different languages/symbols and so forth.  In contrast, even within a single geographical location (i.\,e., the original assumed and unaltered DGP distribution), corruptions can be viewed as low probability samples from the tail of the joint and/or marginal densities.  Even across multiple geographic locations, certain corruptions may remain as low likelihood samples (e.\,g., $P(E_A=e_{snow})$ in Egypt or Florida). Here, a soft intervention on $E_A$, to preferentially emphasize low-probability weather events, allows for the evaluation of robustness for all street signs of interest (i.\,e., by fixing $p^*(E_A)$ and sampling and evaluating over all $C$) yet does not change the true underlying distribution.

While this review is not intended to propose and analyze a new theory, it presents this causal model as a unifying \emph{conceptual} framework and a starting point for establishing a more precise mathematical definition, differentiation, and evaluation of non-adversarial corruption robustness relative to distribution shifts and bounded $L_p$-norm adversarial attacks (i.\,e, interventions on $X$).  Given the lack of rigorous definitions of non-adversarial robustness in the literature prior to this review (discussed next), this causal framework acts as a useful frame of reference in reviewing recent work.

\subsubsection{Robustness Tactics and Evaluations}
To characterize studies in this review, we define a \textbf{robustness tactic} as \textit{an explicit approach toward improving the robustness of a model through a modification of the data, architecture, and/or optimization process}. This differentiates from methods which produce DNNs trained and evaluated on data dominated by high-likelihood samples and which do not make specific claims about robustness to low-likelihood corruptions/alterations. Note that robustness tactics may be referred to as \emph{interventions} in other literature (e.\,g.,~\cite{taori2020measuring}), but we avoid here confusion with the use of \emph{intervention} given our causal perspective of robustness from Sec.~\ref{subsub:robust_defn}.

Additionally, we consider a \textbf{robustness evaluation} to be \textit{a set of experiments which examine a model's performance on data generated from a
counterfactually-altered version of the data generating process in Fig. \ref{fig:scm} that gives preference to certain types of low probability samples from $E$, $S$, $R$, or some combination of these variables.}
A robustness evaluation should make a clear distinction between performance on observed data (obtained from the unaltered data generating process), which gives preference to high-probability samples, and performance on data from the counterfactually altered data generating process, which gives preference to low-probability samples.  This distinction is crucial for demonstrating whether the evaluated method achieves any robustness outside of nominal, high-likelihood conditions.

Lastly, we make this distinction between robustness \emph{tactic} and \emph{evaluation} to allow for the possibility that some studies do not describe an explicit robustness tactic but still demonstrate the implicit robustness of their model through the evaluation (or worse, claim robustness without explicit evaluation).

\section{Related Work}
\label{sec:related_work}
We consider first how non-adversarial robustness has been examined in prior reviews or surveys. To date, robustness in deep learning has been widely studied and reviewed in the context of adversarial machine learning. Several reviews~\cite{thomas2018adversarial, miller2017adversarial, akhtar2018threat, huang2020survey, rawat2017deep} have been conducted in recent years detailing the expansion of adversarial research from the initial discovery in 2014 through the current state-of-the-art advances. 
In recent years, the robustness of deep neural networks to common corruptions and real world transformations has been revisited, first highlighted in~\cite{hendrycks2019benchmarking} and followed up by~\cite{hendrycks2020many}.  Subsequent research results~\cite{taori2020measuring, djolonga2021robustness} have highlighted that neural network performance for more natural forms of image corruption/alteration is limited and poses a serious threat to the deployment of deep learning models in the wild where non-adversarial corruptions are likely to exist. 

\emph{Despite the expanding robustness research beyond adversarial machine learning, to-date there exists no systematic review of neural network robustness to other forms of natural, non-adversarial corruption}. As such, this review intends to identify current tactics and evaluation strategies for measuring neural network robustness.

\section{Methodology}
The protocol for this systematic review was developed according to PRISMA guidelines (preferred reporting items for systematic reviews and meta-analyses)~\cite{Pagen71}.

\subsection{Research Questions}
\label{sub:questions}
This review aims to specifically address the following research questions:

\begin{itemize}
    \itemsep0cm
    \item RQ1 - How is model robustness formally defined in the literature?
    \item RQ2 - What are the datasets and data collection/synthesis strategies used to evaluate non-adversarial robustness?
    \item RQ3 - How is robustness evaluated and measured for conventional computer vision tasks?
    \item RQ4 - What are the primary robustness tactics for computer vision?
\end{itemize}

Answering these questions will provide a deeper understanding of the prevailing definitions of non-adversarial robustness, the extent of the current robustness gap, the tactics towards closing that gap, and the remaining opportunities for future research.

\subsection{Search Strategy}
\label{sub:search}

Relevant studies were identified by searching the Compendex database from January 2012 to July 2021. Compendex contains over 27 million records including those from computer vision and machine learning conferences and journals with considerable relevance to this review. An initial broad search was employed to ensure studies were pulled from all relevant sources.  This search consisted of the following terms

\begin{displayquote}
(robust*) AND (computer vision) AND (deep learning OR deep neural network)
\end{displayquote}

\noindent where the latter two sets of terms were identified as controlled vocabulary to ensure all relative variants (e.\,g., machine vision, deep machine learning, etc.) were also included in the search. Since many approaches to robustness are often described generally and applied/evaluated on computer vision domains, many relevant studies do not explicitly describe vision/perception/etc. in the title and abstract such that the broad search strategy risked excluding these studies.  To address this issue, a separate narrow search was performed whereby the ``computer vision'' terms were dropped and proceedings from the top computer vision and machine learning conferences were specifically targeted.  For more detailed information about this hybrid search strategy, please refer to Appendix \ref{app:search}. Limitations of this approach are discussed later in Sec.~\ref{sec:limitations}.

\subsection{Study Selection}
\label{sub:selection}
Following the removal of duplicates, results from the database searches were first pre-screened using the title and abstract. Obvious exclusions were determined including those which were topic/domain-irrelevant and/or did not describe any robustness considerations. We then proceeded to full-text review, where each study was examined to determine whether a robustness tactic is presented and/or a robustness evaluation performed.  Failure to comply to the described inclusion/exclusion criteria resulted in the study's removal from further consideration. In short, we included studies that present a deep learning-based robustness tactic and/or evaluation for computer vision tasks.  In addition, we excluded studies that focus solely on adversarial robustness or address non-adversarial scenarios in a purely image-processing context. A detailed description of our inclusion/exclusion criteria can be found in Appendix \ref{app:select}.

\subsection{Considerations for Adversarial and Non-Adversarial Examples}
Our definition of robustness in Sec.~\ref{subsub:robust_defn} intentionally excludes adversarial conditions, where perturbations (i.\,e, typically interventions on $X$) are explicitly optimized to fool the model.  We exclude this line of research for the following reasons. First, adversarial machine learning is already a widely-studied and important line of research and warrants a separate robustness review altogether. Several existing reviews already discuss this topic in great detail (see Section~\ref{sec:related_work}).  

Additionally, adversarial examples are \emph{explicitly constructed} (i.\,e, interventions on $X$ for digital attacks or $E$ for patch attacks) to manipulate deep learning model behavior whereas we aim to examine robustness in the context of alterations that occur naturally due to the natural environment and/or sensor and which are entirely independent of the deep learning model and computer vision task (i.\,e, direct samples from the data generating distribution or those with simulated effects consistent with a causal intervention). While adversarial examples still represent an important perspective on worst-case performance, recent results~\cite{laugros2019_are-adversarial} have raised questions about the strength of connections between robustness against synthesized adversarial examples and more naturally occurring corruptions.  

Furthermore, while adversarial machine learning research is able to provide more precise mathematical definitions for image perturbations, these constraints often do not translate to real-world scenarios.  For instance, $L_p$-norm bounds $||\eta||_p \leq \epsilon$ are typical for constraining adversarial perturbations, yet common corruptions often trivially violate these bounds (e.\,g., salt-and-pepper noise - See Appendix \ref{app:lp_bounds}). As such, this complicates the ability to understand and generalize adversarial methods to more common scenarios and further demonstrates the arbitrary and overly-constrained nature of this perturbation-based definition.

Lastly, unless a study under consideration explicitly evaluates non-adversarial corruptions/ alterations, we can only speculate as to whether an adversarial-robustness tactic translates to non-adversarial conditions. As such, we focus explicitly on methods which directly address or evaluate on natural corruptions/perturbations (i.\,e, samples from the tails of the marginals of $E, S, R$) and exclude those with only an adversarial focus.

\subsection{Data Extraction Strategy}
For studies which met the inclusion criteria, we perform a detailed extraction aimed at answering the research questions from Section~\ref{sub:questions}. A data extraction template was developed and is summarized in Appendix \ref{app:extract}.

\subsection{Synthesis of Data}
Data is synthesized and organized according to the four research questions outlined in Section~\ref{sub:questions}. Studies were grouped and compared according to the types of robustness tactics and evaluations performed. General trends were identified across both included and excluded papers while detailed trends were extracted from the included studies only. The results of the overall search are summarized in Figure~\ref{fig:prisma}. Note, for various analyses, we indicate paper counts in parentheses in the text.

\section{Results}
\subsection{General Trends }
\begin{figure*}[t!]
    \centering
    \begin{minipage}{.5\textwidth}
        \centering
        \includegraphics[width=\linewidth]{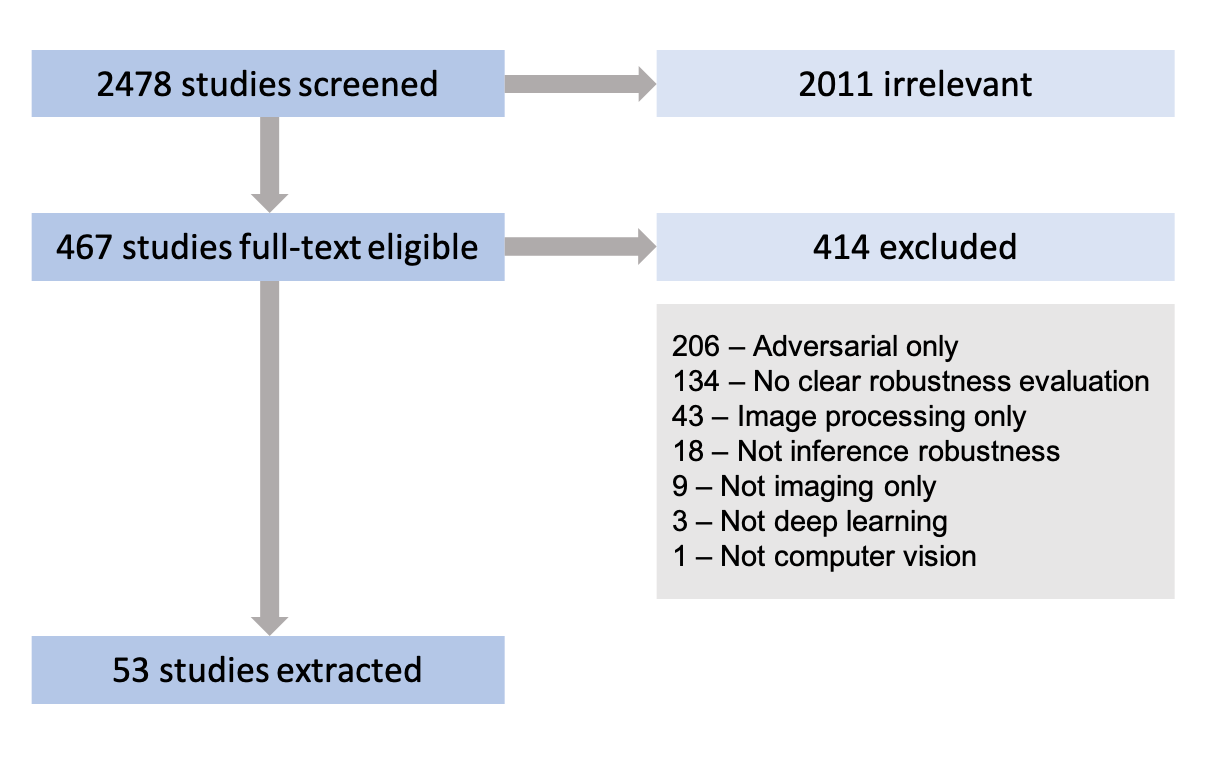}
        \caption{PRISMA diagram}
        \label{fig:prisma}
    \end{minipage}%
    \begin{minipage}{.5\textwidth}
        \centering
        \includegraphics[width=\linewidth]{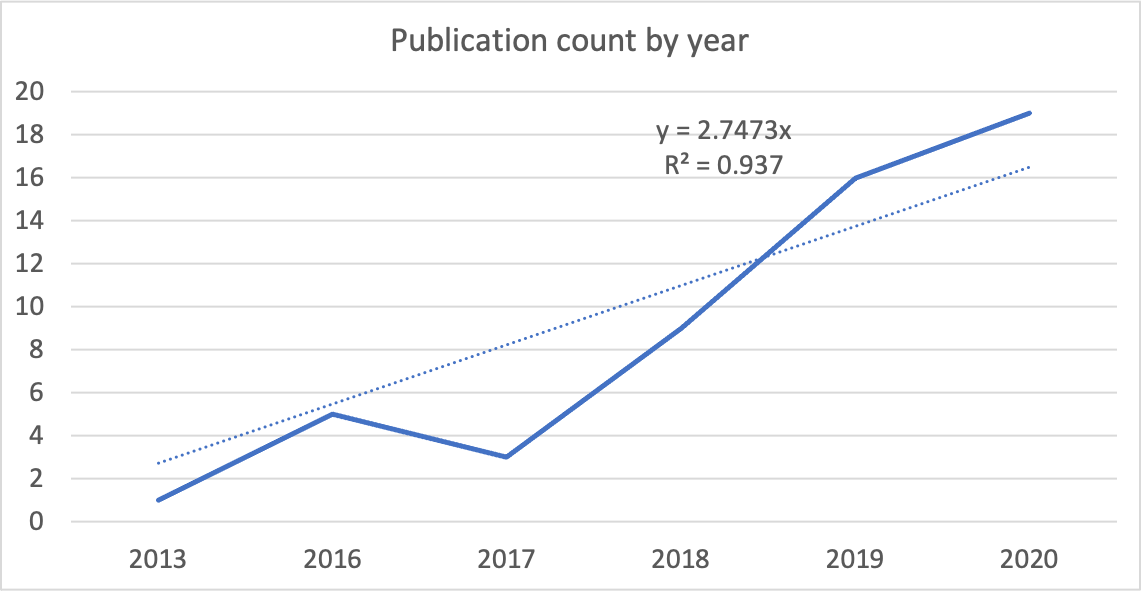}
        \caption{Corruption robustness publications by year}
        \label{fig:pubs_by_year}
    \end{minipage}
\end{figure*}

Included studies (53) related to non-adversarial robustness extend back to 2013 with the number of annual publications increasing steadily in the years since (see Figure~\ref{fig:pubs_by_year}). Publications were spread across well regarded computer vision and machine learning venues including CVPR, NeurIPS, ICCV, ECCV, ICML, MICCAI, and ICIP (among others). In combination, these provide clear evidence that the computer vision and machine learning community are increasingly invested in studying this aspect of deep learning robustness. However, as Figure~\ref{fig:prisma} illustrates, studies on adversarial robustness alone are significantly more widespread and account for nearly 50\% of excluded studies. This fraction is especially interesting given that our search did not explicitly include adversarial-related terms. This further illustrates both the disproportionate focus on adversarial research relative to non-adversarial conditions as well as the confounding robustness terminology.

In terms of motivation, 31 studies focus explicitly on corruption robustness and provide both a robustness tactic and evaluation. These tactics were split relatively evenly across data augmentation (13), architecture (10), and optimization (8). While methods may overlap across these labels, they are assigned according to their primary category. Additionally, 12 studies focus on more general purpose methods (i.\,e, providing no explicit robustness tactic) split over architecture (8) and optimization (4) and address robustness through explicit evaluation. An additional nine studies provide an evaluation only, absent of any novel tactics for addressing the non-adversarial robustness problem.

\subsection{RQ1 - How is model robustness formally defined in the literature?}
Of the 53 studies analyzed, formal mathematical definitions of robustness were largely absent. \cite{zhao2020_maximum-entropy} loosely connects corruption robustness to the single-source domain generalization problem (formally defined in~\cite{qiao2020learning}). \cite{zheng2016_improving} interprets robustness in the context of stability training with the perspective that (\emph{clean, perturbed}) image pairs with small distance between them should similarly have small distance in feature space.  

Of most relevance, \cite{hendrycks2019_benchmarking} defines robustness as average-case performance over a set of corruptions which leads to their definition of mean Corruption Error (mCE), relative Corruption Error (rCE), and Flip Probability (FP). 
The metrics $mCE$ and $rCE$ can be defined as:

\begin{align*}
    mCE &= \sum_{c\in\mathcal{C}}\sum\limits_{s=1}^{5}\frac{E_s^c}{E_{AlexNet, s}^c} \\
    rCE &= \sum_{c\in\mathcal{C}}\sum\limits_{s=1}^{5}\frac{E_s^c - E_{clean}}{E_{AlexNet, s}^c - E_{AlexNet, clean}}
\end{align*}

\noindent where $E$ is the classifier's error rate, $\mathcal{C}$ is a set of corruptions, and $s$ is the severity of the corruption.  The AlexNet model~\cite{krizhevsky2012imagenet} is chosen to provide a common point of reference across DNN models. 

Similarly, \cite{laugros2019_are-adversarial} provides a set of corruption categories and the robustness score to measure the ratio of the model's corruption accuracy to its clean accuracy ($R^\phi_M = \sfrac{A_\phi}{A_{clean}}$, for model $M$, corruption $\phi$, and where $A$ denotes accuracy). Both \cite{hendrycks2019_benchmarking} and \cite{laugros2019_are-adversarial} provide multiple corruption conditions as representative examples (such as varying noise, blur, weather, and rendering conditions) and provide novel metrics as proxies for describing and measuring robustness.  While the corruptions identified in these studies were motivated by real world experiences and prior work, retroactive mapping to the data generating model in Figure~\ref{fig:scm} would enable a more concrete and quantitative analysis regarding their respective likelihoods and effects.

While these metrics incorporate both corruption type and severity, they assume equal likelihood and weighting between these conditions which may not reflect the real world where natural phenomena occur with widely varying frequencies and consequences. The data generating model (Fig.~\ref{fig:scm}) provides a more concrete framework for characterizing or modeling relevant differences in corruption/alteration likelihoods. In simulated datasets, these likelihoods could be determined from known causal graphs.  In real datasets, likelihoods might be estimated from distributions of proxy measurements (e.g., PSNR) or image metadata.

Lastly,~\cite{taori2020measuring} presents two metrics for \textit{effective} and \textit{relative} robustness which considers how performance on natural distribution shifts relates to performance on an unperturbed test set. In our causal framework, this would be akin to determining how accuracy on low-likelihood, corrupted samples from the tail of the distribution compares to accuracy on high-likelihood, uncorrupted images.  The metrics are defined as follows:
\begin{align*}
    \rho(f) &= acc_2(f) - \beta \cdot acc_1(f) \\
    \tau(f') &= acc_2(f') - acc_2(f)
\end{align*}

\noindent where $f$ is the model under test, $acc_1, acc_2$ are the accuracy on the original and shifted datasets respectively, $\beta$ is a log-linear fit to the baseline accuracy of a large number of external models on the unperturbed test set, and $f'$ represents the model resulting from a robustness intervention.  Larger values of these metrics indicate robustness improvements over aggregate and individual models and the effective robustness measure in particular better captures how well a specific model does beyond what is expected given a population of models. However, as with $mCE, rCE$, these metrics do not consider the likelihood or nature of the image corruptions which may be important for estimating how well a model will perform in more realistic settings. 

The general absence of rigorous mathematical definitions of real-world robustness in these studies speaks to the on-going ambiguous and confounding uses of the term in the research community.

\subsection{RQ2 - What are the datasets and data collection/synthesis strategies used to evaluate non-adversarial robustness?}
\label{sub:data}

Table~\ref{tab:corruptions} provides a breakdown of imaging corruptions according to their source in the data generating process and causal model of Fig.~\ref{fig:scm}. Common benchmarks, such as ImageNet-C/P produced by~\cite{hendrycks2019_benchmarking}, provide well-defined test datasets around which many robustness evaluations were performed. Subsets of these datasets were also used to target evaluations to specific components such as sensor noise ($S$) or lighting ($E_A$) conditions only. In many cases, custom benchmarks were also defined where specific corruptions were gathered naturally or generated synthetically. For both common and custom benchmarks, most corruptions were typically generated synthetically via well-defined pre-processing steps which act as direct interventions on the corresponding node(s) of the DAG in Fig.~\ref{fig:scm}. These methods are often easily transferable to similar domains and new datasets.

While many corruptions/alterations are generated synthetically, their true likelihood relative to the data generating model (Figure~\ref{fig:scm}) is often unknown or not examined in these studies.  While the SCM and associated DAG in Fig.~\ref{fig:scm} support arbitrary interventions on the nodes, in order to test a DNN's true robustness to low-likelihood conditions, it is clearly important to ensure that the interventions found in common benchmarks indeed map to conditions represented in the tails of the data distribution. This mapping is made loosely, if at all, in the studies examined and presents an opportunity for the development of more rigorous data characterization methods in the future.

In relation to the causal model of Figure~\ref{fig:scm}, many robustness tactics (13) were evaluated against a combination of environment $E$, sensor $S$, and rendering $R$ corruptions while the vast majority (25) focused specifically on sensor-related corruptions (e.\,g., Gaussian, shot, or impulse noise). Environmental corruptions (typically lighting, rain) received the next most attention (9) while rendering corruptions (e.\,g., compression artifacts) were studied the least (7). 

\begin{figure*}[h!]
    \centering
    \includegraphics[width=0.5\linewidth]{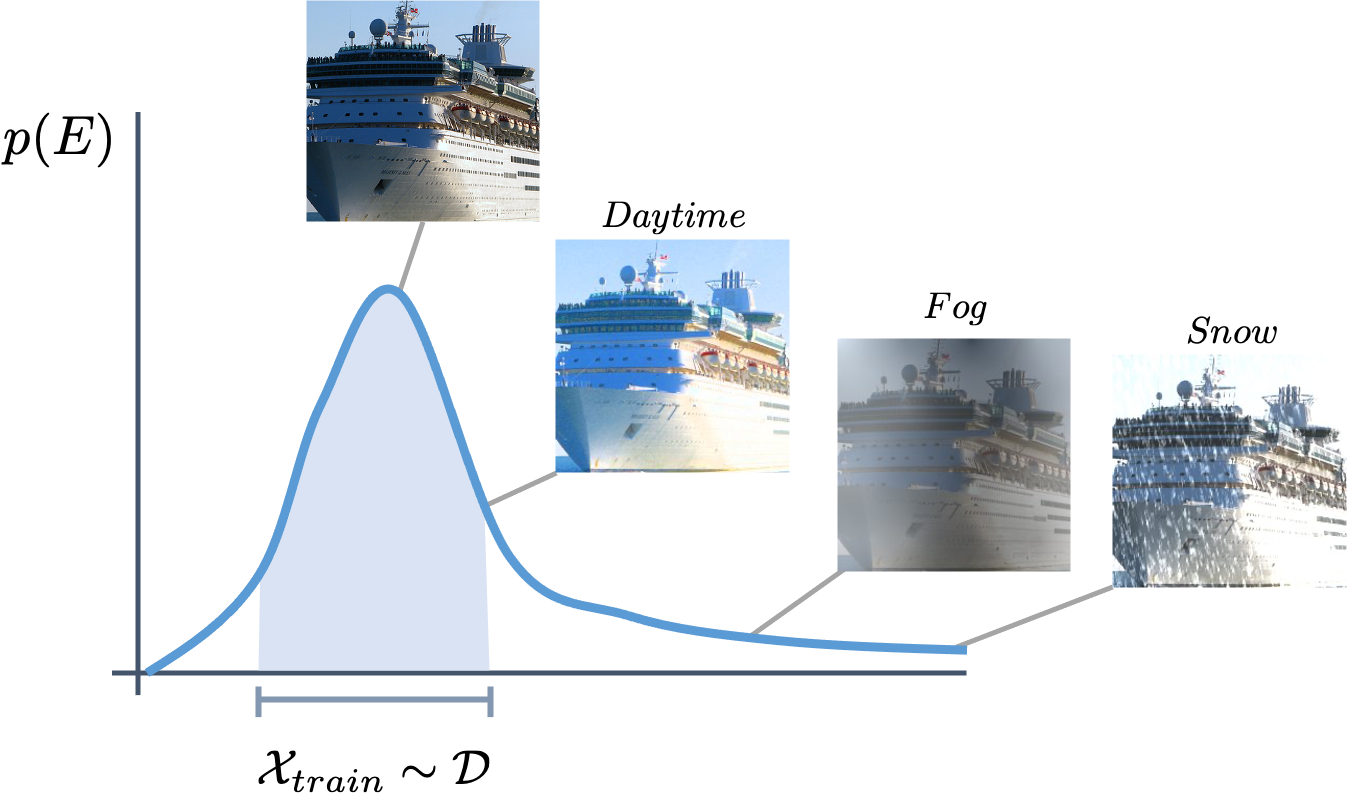}
    \caption{Marginals from the data generating model provide insights into the nature, severity, and likelihood of image alterations/corruptions.  Here shown is an illustration of increasingly unlikely weather events sampled from the marginal distribution for the sensor $(E)$. Note that the training set $\mathcal{X}_{train}$  is most likely sampled from the body of the full distribution given the higher likelihood of those samples and the finite size of $\mathcal{X}_{train}$. Images adapted from ImageNet, ImageNet-C~\cite{hendrycks2019_benchmarking}.}
    \label{fig:my_label}
\end{figure*}

\begin{table*}[]
    \caption{Types of corruptions used during evaluation grouped according to their data source.}
    \resizebox{\linewidth}{!}{
    \centering
    \begin{tabular}{p{2.5cm}|p{4.3cm}|p{2.4cm}|p{2.8cm}}
    \toprule
        {\bf Corruption Type} & {\bf Example Corruptions} & {\bf Common Benchmark} & {\bf Custom Benchmark} \\
    \midrule
        {\bf E}nvironment ($E$)  &  {weather (e.\,g., rain, snow, fog, dusk/dawn), lighting, glare, atmospheric/medium effects, occlusions} & \cite{xie2019_multi-level, xu2019_extended} & \cite{windrim2016_unsupervised, steffens2019_can-exposure, rad2017_alcn:, subramaniam2018_ncc-net:, steffens2020_a-pipelined, li2019_rainflow:, costante2016_exploring} \\
    \midrule
        {\bf S}ensor ($S$) & { noise (e.\,g., Gaussian, impulse, shot), motion blur, zoom blur, defocus blur, lens distortion, color aberration, contrast} & \cite{li2020_wavelet, wu2020_recognizing, xie2019_multi-level} & \cite{geirhos2018_generalisation, nascimento2018_a-robust, chai2018_characterizing, wang2019_learning, li2019_learning, sun2020_implicit, huang2018_some, geirhos2019_imagenet-trained, sun2018_feature, rodner2016_fine-grained, aspandi2019_robust, afifi2019_what, santana-de-nazare2018_color, liu2016_evaluation, alkaddour2020_investigating, steffens2019_can-exposure, tadros2019_assessing, steffens2020_a-pipelined, bastidas2019_channel, jaiswal2020_mute:, costante2016_exploring, dodge2017_can-the-early} \\
    \midrule
        {\bf R}endering ($R$) &  { digital artifacts, filters, JPEG/video compression, sketch/line drawing, cartoonization, quantization} & \cite{wu2020_recognizing} & \cite{zheng2016_improving, wang2019_learning, sun2018_feature, aqqa2019_understanding, tadros2019_assessing, yedla2021_on-the-performance} \\
    \midrule 
        All (\textbf{E, S, R}) & & \cite{dapello2020_simulating, liu2020_how-does, hendrycks2019_benchmarking, zhang2019_making, wang2020_what, zhao2020_maximum-entropy, yang2020_gradaug:, lee2020_smoothmix:, rusak2020_a-simple, li2020_lst-net:, laugros2020_addressing, taori2020measuring} & \cite{laugros2019_are-adversarial, shi2020_informative, taori2020measuring} \\
    
    \end{tabular}
    }
    \label{tab:corruptions}
\vspace{-0.4cm}
\end{table*}

\subsection{RQ3 - How is robustness evaluated and measured for conventional computer vision tasks?}
The vast majority of studies (38) considered robustness in the context of image classification while other standard vision tasks like object detection (4) and segmentation (2) were considerably less popular. Other studies considered optical flow~\cite{harguess2018_an-investigation, li2019_rainflow:}, saliency estimation~\cite{xu2019_extended, sun2020_implicit}, multi-spectral feature learning~\cite{windrim2016_unsupervised}, texture classification~\cite{liu2016_evaluation}, patch matching~\cite{subramaniam2018_ncc-net:}, and visual odometry~\cite{costante2016_exploring}. 

Evaluation procedures explicitly tested baseline models and robustness tactics on corrupted data (as discussed in Sec.~\ref{sub:data}).  In many cases (18), corruptions were also included in the training phase and in a subset of those instances (10), the corruption distribution in the evaluation phase appeared to match that of the training stage which is likely not a fair assessment of a model's robustness to long-tailed events. In principle, since the data generating distribution is expected to be extremely long-tailed, it is unlikely that even prior knowledge about the tails will adequately describe them.  As such, the recommended evaluation procedure would require that DNNs have no prior knowledge of these tails during their optimization and validation stages with their exposure to corruptions occurring the final evaluation. In general, most studies (33) followed a more conventional approach by not augmenting their training set with known/expected corruptions but rather only explicitly addressed them during evaluation.  Since these corruptions are considered long-tail events, it is possible that some may have naturally existed within the training set already.  The key distinction is that these studies did not induce a false distribution shift by over-sampling from the tails during training.

Focusing on the image classification studies, robustness evaluation metrics included standard classification accuracy (28), mean Corruption Error~\cite{hendrycks2019_benchmarking} (7), Robustness Score~\cite{laugros2019_are-adversarial, laugros2020_addressing} (2), and precision at Top-1~\cite{zheng2016_improving}. Of the studies which examined multiple corruption categories (31), noise (20) and blur (5) sensor ($S$) corruptions were found to have the strongest impact on performance. In the noise category, even when employing the robustness tactic, 11 studies exhibited an accuracy degradation (relative to clean accuracy) of more than 20\% (with 6 of those greater than 30\%).  While the observed performance degradation for most corruption categories is not as severe as with adversarial examples, the magnitude of the drop is still significant enough to raise concerns about the use of these models in real world and safety-critical scenarios.

\subsection{RQ4 - What are the primary forms of robustness tactics for computer vision?}
A summary of the papers organized by tactic can be found in Table~\ref{tab:robustness}. Studies were binned according to the primary focus of their approach: (1) architecture, (2) data augmentation, (3) optimization, or (4) evaluation only.

\begin{table*}[h!]
    \caption{Papers organized by robustness tactic}
    \centering
    \resizebox{\linewidth}{!}{
    \begin{tabular}{c|l|l}
    \toprule
    {\small\bf Tactic Type } & {\small\bf Explicit (robustness as motivation)} & {\small\bf Implicit (robustness via evaluation)} \\
    \midrule
    Optimization & \cite{wang2019_learning,  shi2020_informative, li2019_learning, cheng2018_visual, zheng2016_improving, rusak2020_a-simple,  xie2019_multi-level, jaiswal2020_mute:} & \cite{puch2019_few-shot, zhao2020_maximum-entropy, yang2020_gradaug:, nascimento2018_a-robust} \\
    Architecture  & \cite{dapello2020_simulating, liu2020_how-does, li2020_wavelet, wu2020_recognizing,  windrim2016_unsupervised, agostinelli2013_adaptive, sun2018_feature, aspandi2019_robust, rad2017_alcn:,  li2019_rainflow:, costante2016_exploring} & \cite{zhang2019_making, yang2020_interpolation, wang2020_what, subramaniam2018_ncc-net:, xu2019_extended, bastidas2019_channel, li2020_lst-net:} \\
    Data  & \cite{geirhos2018_generalisation, geirhos2019_imagenet-trained, kim2020_puzzle, huang2018_some, zhang2020_corruption-robust, lee2020_smoothmix:, laugros2019_are-adversarial, afifi2019_what, yedla2021_on-the-performance, laugros2020_addressing, harguess2018_an-investigation, santana-de-nazare2018_color, alkaddour2020_investigating} &  \\
    (None) Evaluation only & \multicolumn{2}{l}{\cite{chai2018_characterizing, hendrycks2019_benchmarking, rodner2016_fine-grained, liu2016_evaluation, steffens2019_can-exposure, aqqa2019_understanding, tadros2019_assessing, steffens2020_a-pipelined, dodge2017_can-the-early}} \\
    \end{tabular}
    }
    \label{tab:robustness}
\vspace{-0.4cm}
\end{table*}

\subsubsection{Architecture}
Architectural robustness tactics can be roughly categorized according to the scale of the design elements ranging from single computational layers to design patterns/motifs to full architectures. The scale of the approach speaks to its generalizabilty since a single layer is more likely to have general utility compared to a specialized full network.

In tackling the non-adversarial corruption robustness problem, the majority of architecture-based tactics (18) were found to be full architectures (8) or motifs (4). Several task-specific methods such as~\cite{subramaniam2018_ncc-net:, costante2016_exploring, li2019_rainflow:, xu2019_extended} sought custom architectural solutions for image patch-matching, visual odometry, optical flow, and saliency detection, respectively. While these methods demonstrate some robustness to various conditions, the specificity and diversity of the architectures for each task prevents drawing more generalized conclusions about the true source of robustness in these cases.

In contrast, ~\cite{yang2020_interpolation} introduced the Neural Ordinary Differential Equation (ODE) for interpolating between residual and non-residual neural networks. The inclusion of a damping term in the ODE formulation of~\cite{yang2020_interpolation} ensures local stability of the model to input perturbations. The Neural ODE was further extended in~\cite{liu2020_how-does} by including stochastic network elements such as Dropout, Drop Block, stochastic depth, and random smoothing.  Theoretical analysis points to the added stochastic jump-diffusion terms as providing a guarantee on the asymptotic stability of the ODE. Both methods consider stability in terms of bounded perturbations (akin to adversarial conditions) but also demonstrate strong performance under non-adversarial corruption conditions (2-4\% improvement on CIFAR10-C relative to the ODE variant but a >10\% gap persists between the accuracy on the clean vs. lowest severity conditions).

Motif-based architectural tactics such as~\cite{agostinelli2013_adaptive, aspandi2019_robust, rad2017_alcn:, bastidas2019_channel, dapello2020_simulating, zhang2019_making} provide finer granularity for characterizing robustness gains. On the pre-processing end,~\cite{agostinelli2013_adaptive, aspandi2019_robust} introduce modules for image denoising while \cite{rad2017_alcn:} addresses contrast normalization. 

Taking inspiration from biology,~\cite{dapello2020_simulating} introduces the VOneBlock which consists of a CNN model with fixed-weights modeling known properties of early visual processing in primates.  The module employs fixed-weight Gabor filtering followed by a non-linearity (for distinguishing simple and complex cells) and a stochastic layer. Ablation studies illustrate that removing high spatial frequency Gabor filters and stochasticity both lead to improved corruption robustness (while adversely impacting adversarial robustness). Nonetheless, the reported robustness gap between clean and corrupted images from ImageNet-C is still -31.7\% top-1 accuracy.

A late binding approach, \cite{bastidas2019_channel} uses an attention module to compute a weighting over feature maps from multiple input streams (e.\,g., visible and infrared streams) which allows implicit adaptation to imaging corruptions such as lighting and noise. 

More general purpose computational layers were introduced in \cite{zhang2019_making, li2020_wavelet, li2020_lst-net:}. Both \cite{zhang2019_making} and \cite{li2020_wavelet} take a signal processing perspective in addressing corruption robustness. In particular, \cite{zhang2019_making} returns to first principles by designing anti-aliasing versions of max/average pooling and strided convolutions. In contrast, \cite{li2020_wavelet} provides wavelet-based alternatives to down/up-sampling and demonstrates added robustness to noise corruptions. Lastly, inspired by the Discrete Cosine Transform,~\cite{li2020_lst-net:} developed a series of learnable sparse transforms (LST) with the intent of reducing feature redundancies which can be applied over space, channels, and for importance-based feature map resizing. LST bottlenecks can easily replace bottlenecks in common CNN architectures (e.\,g., ResNet, AlexNet) and are shown to provide accuracy improvements (70.54 mCE vs. 77.01 mCE for LST-Net-50, ResNet-50 respectively) across all corruption categories in ImageNet-C~\cite{hendrycks2019_benchmarking} relative to the unmodified architectures.

The diversity of architectural strategies provides little indication of a clear direction forward. Several methods (\cite{zhang2019_making, li2020_wavelet, dapello2020_simulating}) take a spectral perspective and encourage DNNs to preferentially exploit low-frequency information over high-frequency features. While this does not solve the robustness problem (even in the high-frequency noise scenarios), it presents a minor theme amongst an otherwise scattered set of custom architectures or motifs with limited theoretical grounding and partial empirical success. 

From a causal perspective, we can utilize our knowledge of the data generating process to build robust architectures using a two step approach. By viewing the distributional shift,  noise, or other corruptions as an intervention, as done in \cite{christiansen2021causal}, we can attempt to learn the inverse of the intervention mechanism. Then in the first step, we utilize the inverse function to transform the test image back to better match the distribution of the training images. Then second step involves performing classification as usual on this transformed image.

\subsubsection{Data Augmentation}
Data augmentation tactics aim to increase robustness by diversifying the training dataset to encourage models to be invariant to features or concepts not included in the set $\mathcal{C}$. Recent work~\cite{geirhos2018_generalisation} provided substantial empirical results that demonstrate that conventionally trained deep learning models fail to maintain accuracy when exposed to various forms of noise (e.\,g., uniform, Gaussian, salt-and-pepper) and distortion (e.\,g., Eidolon, contrast, rotation). Their work further illustrated that augmenting the training data with one or more forms of corruption leads to a preservation of accuracy on the seen corruptions but degradation with respect to unseen corruptions outside the training augmentations. Additional studies report consistent findings~\cite{huang2018_some, harguess2018_an-investigation, alkaddour2020_investigating, laugros2019_are-adversarial, yedla2021_on-the-performance}.  

In more specific corruption scenarios, ~\cite{santana-de-nazare2018_color} suggests training on quantized grayscale versions of original RGB training datasets to gain robustness to Gaussian noise corruption. To address color constancy errors, \cite{afifi2019_what} presents a novel approach which combines white balancing as a pre-processing and augmentation step. Their method estimates color transformation matrices using a nearest neighbor search over compressed color features of the training dataset which can be applied to new test images to perform color correction.

Later, Geirhos et al.~\cite{geirhos2019_imagenet-trained} provide substantial evidence that neural networks are over-reliant on texture information which they demonstrate is in stark contrast to humans who exhibit much stronger shape preferences for classification. 
To reduce this dependency, the ImageNet training dataset was augmented using style transfer~\cite{gatys2016image} to create additional texturized variants. Results demonstrated measurable accuracy improvements on the unseen common corruptions in ImageNet-C (69.3 mCE vs. 76.7 for ResNet-50 with and without Stylized ImageNet augmentation) as well as evidence that this form of augmentation increases the shape-bias of DNNs.

Lastly, a new class of methods deriving from Mixup~\cite{zhang2018mixup} have led to steady improvements in corruption robustness. The original formulation used linear mixing of training images and labels 
to encourage more stable predictions on data outside the training distribution (e.\,g., corrupted test images). 
Following~\cite{zhang2018mixup} came related studies such as~\cite{kim2020_puzzle, lee2020_smoothmix:, laugros2020_addressing, zhang2020_corruption-robust}. The PuzzleMix method~\cite{kim2020_puzzle} extends standard Mixup by optimizing the degree of information retained by original pairs of images as well as performing optimal transport between the two images to preserve saliency information in the merged result. Smoothmix~\cite{lee2020_smoothmix:} was developed to alleviate strong edge effects as might be observed in augmentation methods like PuzzleMix. Whenever a candidate pair of images 
is sampled for mixing, Smoothmix generates a smooth blending mask determined by first randomly sampling the mask shape and then associated shape parameters (e.\,g., square or circle masks with sampled dimensions). Images are combined using the mask and labels are smoothed using the relative proportions of each image derived from the mask. Reported gains amount to 1.03\% error improvement on CIFAR100-C but a -31.4\% gap relative to the error for the clean CIFAR100 dataset. Later methods like AugMix~\cite{hendrycks2019augmix} further improved corruption robustness  by mixing with multiple transformations and augmenting the training loss with a Jensen-Shannon divergence between predictions on clean and augmented data points. The method was shown to transfer well to other domains such as in Peripheral Blood Smear Images~\cite{zhang2020_corruption-robust}.

Data augmentation has proven a popular research strategy and achieves moderate empirical success albeit often lacking theoretical grounding. From the causal perspective discussed earlier, data augmentation can be thought of as performing a soft intervention on the distribution of the covariates, so as to generate more low-likelihood samples and thus allow models to more effectively learn relevant features from those samples (a perspective consistent with~\cite{zhang2018mixup, kim2020_puzzle, lee2020_smoothmix:, laugros2020_addressing, zhang2020_corruption-robust}). While these data augmentation tactics certainly increase the diversity of the training samples processed by the DNN, there is still no guarantee that such samples are in the true tail of the data generating model's distribution.  Furthermore, even in its high-level description, each node of the data generating model in Fig.~\ref{fig:scm} is intrinsically complex and may be further factorized into additional sub-graphs, ultimately yielding complex distributions where relying on data augmentations is unlikely to scale effectively (as shown more directly in~\cite{geirhos2018_generalisation}). Using the framework of selection bias, additional tools from that literature could be borrowed and adapted to better address these data-related challenges.

\subsubsection{Optimization}
Optimization-based robustness tactics modify learning objectives to improve resilience to corrupted inputs and encourage learning informative/discriminative features. The design of the loss function or training process attempts to remove potential learning shortcuts which lead to poor generalization and brittleness in the model.

Recently, adversarial training has emerged as a class of techniques which bridges the gap between pure data augmentation and optimization approaches. Methods such as~\cite{madry2018towards} construct a minmax problem where the inner maximization aims to find effective perturbations $\delta$ from some distribution $\Delta$ (e.\,g., adversarial or noise) while the outer minimization aims to update the model parameters $\theta$ to reduce expected error. Most closely related to this approach is the work by~\cite{rusak2020_a-simple} which learns a noise distribution from which to sample $\delta$. This Adversarial Noise Training approach yields an 11.2\% improvement in Top-1 accuracy on ImageNet-C (relative to a standard trained ResNet-50 baseline) but a -25.7\% gap between the clean and corruption accuracy remains (76.1\% vs. 50.4\%).

A novel information theoretic approach is taken in~\cite{zhao2020_maximum-entropy} which uses the maximization phase to find perturbations which simulate ``hard'' target domains. Using the Information Bottleneck principle, the inner phase is regularized by a maximum-entropy term which acts as a lower bound on the mutual information between the input and latent representations. This subsequently encourages perturbations which are not easily compressed and forces the optimization to be selective in the learned representations (ideally resulting in more corruption-resistant features).

Separately,~\cite{wang2019_learning} reformulate the adversarial training so that the maximization aims to reduce the label predictability given spatially local features while the minimization is used to train an auxiliary classifier to be highly predictive on the same local features.

Beyond adversarial training, regularization techniques add terms to the learning objective to further constrain the optimization. Leveraging neural recordings from mice,~\cite{li2019_learning} introduces a form of regularization which encourages similarity between learned representations and measured brain activity which led to increased robustness against additive noise (a 30\% accuracy gain at the highest noise level relative to the unmodified baseline but a >30\% accuracy drop relative to the noiseless case for the neural-regularized model). In contrast, ~\cite{yang2020_gradaug:} uses losses computed over sampled sub-networks which to further regularize the optimization. Unlike standard Dropout, structured sampling is performed by leveraging the model's architecture when computing gradients based on the regularization terms. Another structured dropout strategy is employed by~\cite{shi2020_informative} where output neurons are dropped at rates proportional to the self-information of the corresponding input image patches. Additionally, \cite{nascimento2018_a-robust} uses a sparse coding approach which aims to minimize the $L_0$ norm (or a relaxation thereof) of the output features.

Via contrastive learning, ~\cite{zheng2016_improving} adds a stability loss term based on pairs of clean, perturbed image features. ~\cite{puch2019_few-shot} uses a triple loss to learn an embedding space more strongly tied to discriminative image features while avoiding learning spurious correlations. Using multiple domains, ~\cite{cheng2018_visual} provides a tactic for pre-training using multiple degradations whereas~\cite{xie2019_multi-level} leverages gradients from the loss computed while training an auxiliary domain classifier to learn domain-invariant features. Lastly, \cite{jaiswal2020_mute:} present MUTE, a novel approach which moves away from standard one-hot encodings in classification tasks with a method for finding alternative target encodings which capture inherent ambiguities in the training sample.

While these optimization methods (e.\,g., adversarial training, contrastive methods) also manipulate input data (akin to augmentation), they do so in concert with a specific objective (rather than agnostic to it).  
These optimization techniques can be seen as attempts to learn the invariant causal mechanism corresponding to the function $\kappa$ in Fig.~\ref{fig:scm}. However, by explicitly positing a data generating process and identifying the causal parents of $Y$, these techniques could further target the optimization toward learning the invariant mechanism that will possess the desired robustness properties.  

However, this connection to augmentation suggests that these approaches are still attempting to model sampling from the tails of the data-generating distribution during training rather than encouraging the model to learn the true underlying concepts. Since the data manipulations are not necessarily guaranteed to accurately simulate or come from the true tails of the data distribution, there are currently no guarantees that they will also lead to improvements in robustness equally across all corruption conditions.  Reported tactics demonstrate measurable improvements relative to baseline methods (often >10\%) but in most cases a robustness gap persists (>20\% difference in clean/corrupted accuracy).

\section{Limitations}
\label{sec:limitations}
In order to provide a reasonable scope to this review, the search terms, sources, and assessment criteria we used may have resulted in the exclusion of potentially relevant papers. In particular, by excluding adversarial machine learning papers (unless explicitly evaluated outside the synthesized attack paradigm) this review may ignore a non-negligible number of methods which are capable of translating adversarial robustness to non-adversarial robustness challenges. 

Additionally, while search terms were chosen judiciously, the final set still imposes unavoidable bias on the search results. The use of controlled vocabulary in the search should improve the quality of the returned matches but may introduce biases/inconsistencies depending on the accuracy of the indexing methodology employed by the electronic database. 

Furthermore, as described in Section~\ref{sub:search}, general and targeted searches were performed to limit the total number of papers to a reasonable quantity for the initial screening. This naturally increases the likelihood that relevant papers were excluded. However, the search terms and range of conference venues were sufficiently broad to ensure that relevant papers would not be missed. While many studies in this domain are often released as pre-prints (e.\,g., via sites like arXiv), our exclusion of non-peer reviewed studies is likely to further bias the final pool of studies in our analysis. However, the peer-reviewed publications included in our search still enable us to identify key themes and consistencies in quantitative results.  Outside of the results from our systematic search, a small number relevant papers were identified to provide additional context as necessary but which remain consistent with the main themes of the review.

\section{Discussion and Perspectives}
\subsection{Formalizing Robustness}
The results of this review highlight several trends and shortcomings in the field. First and foremost, the mathematical definition of robustness remains largely missing from on-going research. While several studies attempted partial definitions, the working assumption in most studies suggests simply that models should not sacrifice performance in the face of altered/corrupted inputs.  However, these partial definitions fail to specifically address: 

\begin{itemize}
    \item Under what real world conditions should this performance constancy assumption hold?
    \item What kinds of alterations/corruptions are valid under this assumption? 
    \item How does the severity/likelihood of image alterations/corruptions matter? 
    \item To what extent does the clean accuracy of the model influence the determination of a model's `robustness'? 
    \item What is the role of the train and test sample distributions in this context? 
    \item What is the likelihood of the alteration/corruption and what is the effect of this likelihood on robustness?
    \item How should robustness evaluation and measurement account for downstream consequences?
\end{itemize}

In particular, these questions probe at developing approaches to clearly identify the corruption preconditions under which a model's performance is not expected to change. At one extreme, adversarial examples provide worst case corruptions along with a well-defined expectation that models should remain performant under small, bounded perturbations. On the other extreme, heavy noise or blurring can render images unrecognizable (even to humans), suggesting the importance for defining corruption-conditional measures of severity.  
Robustness definitions need to better account for expected model behavior when operating at points between these two extremes.

Furthermore, robustness to non-adversarial corruptions, as described in the literature, is often conflated with robustness to adversarial examples, distribution shift, and domain generalization. Because these are not equivalent, more precise terminology and mathematical descriptions are necessary to disambiguate these cases.  Our data generating model in Section~\ref{subsub:robust_defn} provides an initial attempt at addressing this issue by distinguishing corruptions as low likelihood samples and distribution shifts as changes to particular marginals of the data generating distribution.

On the evaluation end, metrics like $mCE, rCE$ and Robustness Score provide a foundation for studying the effects of clean accuracy and corruption severity on model robustness but are not a substitute for a formal definition. These metrics implicitly mask the rate of change in performance degradation over corruption severities and types. The use of AlexNet~\cite{krizhevsky2012imagenet} in $mCE$ and $rCE$ as a fixed point of reference is essentially arbitrary and severely limits the interpretability of the metric. Future research can build off of these metrics while also considering the questions raised earlier, particularly with respect to incorporating the unequal impact or likelihood of corruptions and the underlying assumptions regarding the DNN under test. 

\subsection{Robustness Tactics}
\label{sub:tactics}
Amongst the studies in this review, robustness tactics were relatively evenly spread across architecture, data, and optimization categories.  Architectural approaches focused heavily on feedforward, convolutional methods.  Methods focusing on full architectures limit their generalizability to new domains and/or corruption scenarios. Methods which introduced smaller motifs offer greater generalizability but often failed to close the robustness gap (especially in the noise case).  On either end of the spectrum, no method was able to demonstrate constant or near-constant performance over all corruptions/alterations and severities and few or no guarantees on performance were provided in the context of non-adversarial distortions.   

Data augmentation techniques, especially those deriving from the Mixup method, yield marked improvements in task performance. Nonetheless, the Mixup-style approaches often require significant additional computational resources while measurable performance degradation still persists. Several studies demonstrated that training on narrow sets of corruption conditions produced narrow robustness for similar conditions, showing that DNNs can benefit from controlling for covariates during training but that the set of covariates is still small compared to the combinatorial explosion for covering all possibilities. Extending these approaches to control for a wider variety of corruptions is simply not practical given the extreme variability of the real world described by the long-tailed distributions from the SCM in Fig.~\ref{fig:scm}.   
Generally speaking, viewing non-adversarial corruptions/alterations as resulting from low-likelihood samples from the environment, sensor, and render nodes ($E, S, R$) of the data generating distribution suggests  augmentation techniques are unlikely to sufficiently scale to capture the tails of the underlying true distribution. Many of the optimization strategies presented take a similar approach and as such, are likely to hit the same limitations. 

Additionally, the growing over-reliance on increasing quantities of training data speaks to the question of whether definitions of robustness should also account for the training sample complexity. Recent work~\cite{geirhos2021partial} suggests that increasing dataset size is currently the most successful strategy towards improving DNN robustness yet, even for datasets of hundreds of millions of images, a measurable gap remains.  If the required sample complexity must continue to grow even beyond these already massive datasets, it raises the issue of whether this strategy is sustainable. Even if the factors of the data generating model could be represented by a reasonably small number of variables, the number of training samples required to ensure sufficient coverage of the domain and generalization beyond training conditions would be prohibitively large.  This suggests current architectures and optimization tactics need to shift accordingly.  Lifelong or curriculum learning offer potential alternative tactics for increasing training data efficiency while dynamic models with recurrence or feedback might enable DNNs to be adaptive to long-tail events and changing conditions.
Future tactics should consider how DNN robustness to non-adversarial corruptions/alterations can be improved without simply increasing the sample complexity.

\subsection{Causal Approaches to Robustness} 
While our SCM and DAG for the image generating process directly addresses robustness as relating to long-tail events, other work\cite{parascandolo2018learning, christiansen2021causal} provides useful alternate causal viewpoints which harmonize well with our interpretation yet offer separate perspectives. 

In these other causal methods, covariate shift and domain adaptation are then viewed as a result of interventions on the data generating process. This allows for robustness to be defined as the ability to predict well when the test data is generated from an interventional distribution.
To learn robust prediction algorithms, these methods utilize the causal invariance principle which states that the causal mechanisms that generate the data are modular, i.\,e., one mechanism does not change when another is intervened on. This means the causal mechanisms are robust to changes in the distribution of the covariates. 

And so \cite{christiansen2021causal} applies this  principle to learn causal regressions -- regressing the response variable $Y$ onto its causal parents. This causal regression possesses robustness properties, and is shown to be optimal in certain settings. Although not explicitly applied in the image domain, the framework is sufficiently general to deal with high dimensional data. 

In a slightly different approach, \cite{parascandolo2018learning} learn the inverse of the mechanism that induces the covariate shift. This is done using a mixture of experts in an unsupervised fashion. Generalizability is achieved by ``inverting'' the shifted data points back to their original distribution and then performing prediction as usual.

This separate line of work offers additional causal interpretations which provide  evidence that while tactics discussed in this review may emulate aspects of these causal processes, more direct application of these techniques could offer greater potential for robustness gains.

\subsection{Conclusions and Future Research Directions}
\label{sub:future}
While much progress has been made in improving the performance of deep learning models on altered or corrupted inputs, the results of this study and separate large-scale empirical efforts~\cite{taori2020measuring, djolonga2021robustness} demonstrate a persistent robustness gap between what is currently achievable and what is expected and necessary for safety-critical applications.  In order to close this gap, an essential consideration for future research is a clear, concise definition of robustness in computer vision which better establishes preconditions such as the expected corruption types, likelihood, and severity, as well as a clear definition of the expected model performance under these conditions which may be based on human performance, reference models, or application-specific requirements. 

In the context of tactics, adaptive models were largely absent in the studies in this review.  For instance, models utilizing top-down or recurrent feedback might offer greater capacity for adaptation to corruptions during the inference stage. The majority of tactics (architectural, data, or optimization-focused) operate on the assumption that a static set of weights can be learned, which is robust in the face of diverse corruption types and severities. This assumption may need to be more heavily challenged in order to achieve greater corruption robustness.

The growing popularity in common corruption robustness research has been well supported by the introduction of new benchmark datasets and metrics~\cite{hendrycks2019_benchmarking, zhang2020_corruption-robust, Altindis2021-uc, Michaelis2019-iw, taori2020measuring} . While these are indispensable resources, future research should take care not to collectively over-fit methods to these benchmarks but should introduce additional custom evaluations when possible in an effort to perform more targeted interventions on the data generating process consistent with the task and expectations about the model.  At a minimum, robustness evaluations should be a standard consideration when introducing novel vision models or algorithms, sometimes requiring custom datasets to probe model performance boundaries.  Furthermore, since it is well-established that DNNs are capable of achieving state-of-the-art performance on a range of tasks, the robustness of these models to non-adversarial alterations/distortions should be considered in the initial design rather than as an afterthought during evaluation. In viewing robustness as a relative measure of performance between events sampled from the tail and body of the data distribution, well-defined expectations about DNN performance for specific interventions on the data generating model should be defined a priori. This way, DNNs can claim robustness under specific conditions when they meet or exceed expectations. Model robustness in this context then stands to act as a differentiator when comparing multiple approaches (rather than simply focusing on state-of-the-art performance).

Future research should strive to develop new methods for better characterizing the properties of training/test distributions including the extent to which the samples adequately represent the tails of the underlying data generating distribution.  While data augmentation and large-scale training datasets are likely not sustainable solutions, one approach to counter these trends would be to develop methods for more intelligently sampling training data. The proposed SCM and DAG in Fig.~\ref{fig:scm} provides a more explicit mechanism for understanding and performing this more targeted sampling. New methods could establish sufficient conditions for datasets with respect to both concept learning and robustness to expected corruptions/alterations. Additionally, given the apparent importance of dataset size and diversity for model robustness, future research could develop novel methods for predicting/estimating a model's robustness as a function of its underlying training dataset.

While adversarial robustness remains a popular and relevant direction of research, the studies in this review demonstrate that natural image corruptions alone yield a significant performance gap. If the research community is content to continue to search for, e.\,g., 1-3\% improvements in performance on clean benchmark tasks, it should be equally invested in protecting those gains against even larger degradation, e.\,.g., 10-30\%, due to common corruptions and other natural, non-adversarial image alterations.  Failure to mind this gap could have severe consequences in future safety-critical applications.

{\small
\bibliographystyle{ieee_fullname}
\bibliography{egbib}
}

\clearpage
\appendix

\section{Search Strategy} 
\label{app:search}
Due to the broad and ambiguous definition of \emph{robustness} in deep learning, a multi-faceted search strategy was necessary to gather relevant studies while managing the total scope of the review.  Broad and narrow searches were conducted using the following search terms.

\noindent Broad-search terms:
\begin{displayquote}
((((robust*) WN All fields) AND ((computer vision) WN CV) ) AND (({deep learning} OR {deep neural networks} OR {neural networks} OR {convolutional neural networks}) WN CV) )
\end{displayquote}

\noindent Narrow-search terms:
\begin{displayquote}
((((robust*) WN All fields) AND (({deep learning} OR {neural networks} OR {convolutional neural networks} OR {deep neural networks}) WN CV) ) AND (("NeurIPS" OR "NIPS" OR "CVPR" OR "ICCV" OR "IJCV" OR "WACV" OR "ECCV" OR "BMVC" OR "ACCV" OR "MVA" OR "ICML" OR "ICLR" OR "MICCAI" OR "AAAI" OR "IROS" OR "ICPR" OR "TPAMI" OR "ICIP") WN CF))
\end{displayquote}

\section{Detailed Study Selection Criteria}
\label{app:select}
\noindent Studies were eligible for our review based on the following criteria:
\begin{itemize}
    \itemsep0em
    \item Deep learning-based methods
    \item Studies which describe a specific robustness tactic
    \item Studies which perform an explicit robustness evaluation
    \item General purpose methods applied to 2D computer vision tasks including, but not limited to, image classification, object detection, tracking, or segmentation.
    \item Studies which focus on inference-time robustness (as opposed to robustness to train-time label noise, backdoor attacks, etc.)
\end{itemize}

\noindent Studies were considered ineligible for this review if they met all of the following exclusion criteria:
\begin{itemize}
    \itemsep0em
    \item Studies which do not describe a robustness tactic 
    \item Studies which do not explicitly evaluate on corrupted/altered data
    \item Studies which do not provide any quantitative results (e.g., surveys, reviews, or opinion papers)
    \item Studies which focus on non-image modalities (e.g., text, audio, point clouds) with no direct evaluation on 2D image data
    \item Studies involving adversarial attacks/defenses without analysis of robustness to non-targeted corruptions or transformations 
    \item Studies which present deep learning methods for performing image processing on corrupted images including, but not limited to, image denoising, super-resoultion, reconstruction, inpainting, etc.
\end{itemize}

\section{Data Extraction}
\label{app:extract}
Table~\ref{tab:extraction} describes in detail our data extraction approach for the studies included in this review.

\begin{table*}[h!]
    \caption{Data extraction strategy}
    \centering
    \small
    \resizebox{0.8\linewidth}{!}{
    \begin{tabular}{p{2cm}|p{3cm}|p{8cm}}
    \hline
    \hline
       \textbf{Category}  & \textbf{Item} & \textbf{Description} \\
       \multirow{2}{*}{General} & Year & Year of publication \\
       & Venue & Name of peer-reviewed conference or journal \\
       \midrule
        \multirow{3}{*}{Robustness} & Definition & Is a formal definition of robustness provided? \\
        & Origin & (1) Explicit - Robustness is the primary motivation (2) Implicit - Robustness demonstrated via evaluation \\
        & Tactic & (1) Architecture, (2) Data, (3) Optimization, (4) Other, (5) None (evaluation only) \\
        
    \midrule
        Architecture & Strategy & Identify the approach such as pre-processing modules, novel layers, architectural motifs, or full architecture \\
    \midrule
        \multirow{3}{*}{Data} & Training strategy & Primary data strategy (e.g., augmentation, denoising, etc.) \\
        & Training corruptions & Inclusion of corruptions during training \\
        & Train/test alignment & Do the train/test corruption distributions match? \\
    \midrule
        \multirow{2}{*}{Optimization} & Strategy & Optimization approach such as adversarial training, self-supervised learning, contrastive learning \\
        & Loss/objective & Describe optimization loss function \\
    \midrule
        \multirow{6}{*}{Evaluation} & Computer vision task & Primary computer vision task such as classification, detection, optical flow, stereo depth estimation \\
        & Data domain & Types of evaluated images (e.g., natural, remote sensing, medical) \\
        & Benchmark type & Origin of benchmark dataset (e.g., public domain, synthetic, analysis/filtering of existing data) \\
        & Benchmark dataset & Name of benchmark dataset \\
        & Corruptions & Categories of corruptions evaluated (e.g., noise, weather, blur, synthetic) \\
        & Metrics & Robustness evaluation metrics (e.g., accuracy, mCE, MSE) \\
        & Strongest corruption & Category of corruption causing largest degradation in performance \\
        & Pre-tactic degradation & Estimate of performance degradation prior to robustness tactic \\
        & Post-tactic degradation & Estimate of performance degradation prior to robustness tactic \\
    \bottomrule
    \end{tabular}
    }
    \label{tab:extraction}
\end{table*}

\pagebreak

\section{Violation of Perturbation Bounds for Non-adversarial Conditions}
\label{app:lp_bounds}

In the case of salt-and-pepper noise, we can write the corrupted image as $x' = clip(x + \eta, 0, 1)$ where $\{x, x', \eta\} \in \mathbb{R}^{C\times W \times H}$ and

\[
\eta_{i,j} =
\begin{cases}
    -1 & \text{with probability } p \\
    1 & \text{with probability } q \\
    0 & \text{with probability } 1 - (p+q)
\end{cases}
\]

If $\eta_{i,j}$ is non-zero, then it either fully nulls or saturates the value at that location. The $L_p$-norm constraint is not useful in this case and illustrates the need to further consider the appropriate form of constraint on the robustness-related interventions using the data generating model from Figure 1. This is particularly important given that $L_p$-norm constraints (or similar) are more commonly examined in the machine learning community in the context of adversarial attack, yet salt-and-pepper noise (or similar) is more likely to occur in the real world.

\end{document}